# A New ECOC Algorithm for Multiclass Microarray Data Classification

Mengxin Sun; Kunhong Liu(✉); Qingqi Hong; Beizhan Wang
Software School of Xiamen University, Xiamen, China
mengxinsun.xmu@gmail.com; lkhqz@xmu.edu.cn; hongqq@xmu.edu.cn;

*Abstract*—The classification of multi-class microarray datasets is a hard task because of the small samples size in each class and the heavy overlaps among classes. To effectively solve these problems, we propose novel Error Correcting Output Code (ECOC) algorithm by Enhance Class Separability related Data Complexity measures during encoding process, named as ECOCECS. In this algorithm, two nearest neighbor related DC measures are deployed to extract the intrinsic overlapping information from microarray data. Our ECOC algorithm aims to search an optimal class split scheme by minimizing these measures. The class splitting process ends when each class is separated from others, and then the class assignment scheme is mapped as a coding matrix. Experiments are carried out on five microarray datasets, and results demonstrate the effectiveness and robustness of our method in comparison with six state-of-art ECOC methods. In short, our work confirm the probability of applying DC to ECOC framework.

## I. INTRODUCTION

DNA microarray is a widely deployed technology for cancer diagnosis, and it has gone from obscurity to being almost ubiquitous in biological and medical research. However, a challenge for researchers is raised due to the "large dimensions, small samples" problem embedded in microarray data analysis: the number of samples is far smaller than that of genes. Because of the severe data unbalanced problem among classes, multi-class microarray data is much harder compared with the binary class problem, and a lot of algorithms were proposed to mine key genes and obtain accurate cancer diagnosis [1].

A popular solution is to pick up importance features to reduce problem difficulty and to complete multi-class classification by Error Correcting Output Code (ECOC) method. As it provides high error correcting ability, it has been successfully applied in many fields, such as face recognition[2], biological disease diagnosis[3] and text recognition[4]. In general, ECOC consists of two main phases: encoding phase and decoding phase. In the encoding phase, a given data set with R classes $\{c_1, c_2, \ldots, c_R\}$, is divided into L binary problems. These separation schemes are represented as a coding matrix M, where $M \in \{+1, -1\}^{R*L}$. Each row in M defines a unique codeword for each class in R, and each column defines a partition of classes with +1, -1 as their classes membership. In the decoding phase, the outputs of L classifiers compose a code vector, $\{L_{1x}, L_{2x}, \ldots, L_{Rx}\}$ for an unknown sample x. This vector is compared with each codeword of M, and the class represented by the codeword with the highest similarity is selected as predicted class, shown as Fig.1(a).

There are two types of ECOC algorithms: data-independent [5] and data-dependent. The difference between them is whether a coding matrix is generated based on data characteristics. Some data-dependent algorithms deploy different measures for data characteristics evaluation, and a typical one is mutual information (MI) in DECOC[3]. However, such measures require a large number of samples to estimate key parameters, so they unavoidably lead to the bias in microarray data due to the small-sample size, degrading their performance [6].

Instead, this paper aims to design a reliable data-dependent ECOC algorithm to tackle the high dimensional microarray data based on Data complexity (DC), which is a powerful tool to explore data distribution and relationships among characteristics. In [7], some complexity measures are defined for binary classification, focusing on the geometrical complexity of the class boundary. Since the classification results are affected by the complexity in a data set, [8] used DC measures to determine the domain of competition among the classifiers. Other researchers also used different complexity measures to choose some promising classification methods[9-11]. In short, DC is helpful in a classification task.

In this paper, we propose a new ECOC algorithm based on Enhancing Class Separability related DC (N2 and N3), named as ECOCECS. It aims to search optimal schemes to split classes into two groups by minimizing N2/N3 indices between groups, so as to improve dichotomizers' performances. Our experiments are based on five microarray datasets, and six widely used ECOC algorithms are employed for comparisons. Experimental results prove our ECOCECS is more accurate and stable.

The structure of paper is as follows. Section 2 briefly introduces N2 and N3, and Section 3 presents the details of ECOCECS. Section 4 analyzes and discusses the experimental results, and Section 5 concludes this paper.

## II. CLASS SEPARABILITY BASED DC MEASURES

### A. Ratio of nearest neighbor distance (noted as N2)

Assume there are N samples in a data set. N2 measure is proposed to judge whether samples in the same class are close or not. The Euclidean distance is calculated from sample $x_i$ to its nearest neighbor inside the class (denoted as $intraDis(x_i)$) and outside the class (denoted as $interDis(x_i)$). Then N2 index is set as the average distance of intra-class nearest neighbor over the average distance of inter-class nearest neighbor, as shown in formula(1). It changes in the range of $[0, +\infty)$. A low N2 index suggests that the samples of the same class lay closely, and a large one indicates a disperse distribution.

$$N2 = \frac{\sum_{i=1}^{N} intraDis(x_i)}{\sum_{i=1}^{N} interDis(x_i)} \quad (1)$$

### B. Loss of the nearest neighbor classifier (noted as N3)

N3 is defined as the loss of neighbor classifier on training data. It is calculated as the mismatching between the predicted label $f(x_i)$ and true label $y_i$ for sample $x_i$, as shown in the formulas (2-3). In the consideration of the class-imbalance and small sample size problem in training data sets, leave-one-out method is used to get robust results.

$$L(y_i, x_i) = \begin{Bmatrix} 1, y_i = f(x_i) \\ 0, y_i \neq f(x_i) \end{Bmatrix} \quad (2)$$

$$N3 = \sum_{i=1}^{N} L(y_i, x_i) \quad (3)$$

This measure shows the compactness of samples in a data set. The domain of N3 metric is in the range of [0,1], and a low index shows there is a large margin in the class boundary.

## III. THE WORKFLOW OF ECOCECS

In our algorithm, a search algorithm is designed to exchange classes in groups iteratively, aiming to search an optimal class splitting scheme with high class separability. Because N2 and N3 are proposed on different principles, they are applied in our algorithm individually.

Let the mean value of all samples ($Cen_{k,i}$) in class $c_k$ represent the centroid of $c_k$ in $G_i$. Function $Dis(a,b)$ represents the Euclidean distance between $a$ and $b$. The class assignment scheme for a group is a column in a coding matrix, directly affects the generalization ability of our ECOC algorithm. The workflow of our algorithm can be summarized as: 1. split groups at random at the first step; 2. evaluates the complexity between based on N2 or N3 measures in current group; 3. search a better splitting scheme to lower the complexity by exchanging the most complex classes in two groups.

N2 index measures the ratio of inner-class distance and inter-class distance. Based on this idea, formula (4) is designed to calculate the inner-group ($Dis(Cen_{k,i}, Cen_{l,i}), k \neq l$) and inter-group ($Dis(Cen_{k,i}, Cen_{h,j}), k \neq h, i \neq j$) distance for each class. For $c_k$ in $G_i$, a small $D_{k,i}$ indicates that $c_k$ is far away from other classes in the same group and close the classes in another group. So the class with the minimum $D_{k,i}$ is considered as the most complex class in the i-th group. By exchanging such classes between two groups, it is expected that the complexity would be lower.

As for N3, the class with the longest distance from other classes in same group is treated as the most complex class. So formula (5) calculates the distance between each class pair in a group. Then the class far away from other classes in its group is to be exchanged. The detailed working process is given as Fig.2.

$$D_{k,i} = \frac{\sum_{\forall c_k \in G_i, c_l \in G_i, k \neq l} Dis(Cen_{k,i}, Cen_{l,i})}{\sum_{\forall c_k \in G_i, c_h \in G_j} Dis(Cen_{k,i}, Cen_{h,j})}, i, j \in [1,2], i \neq j \quad (4)$$

$$D_{k,i} = \sum_{\forall c_k \in G_i, c_l \in G_i, k \neq l} Dis(Cen_{k,i}, Cen_{l,i}), i \in [1,2] \quad (5)$$

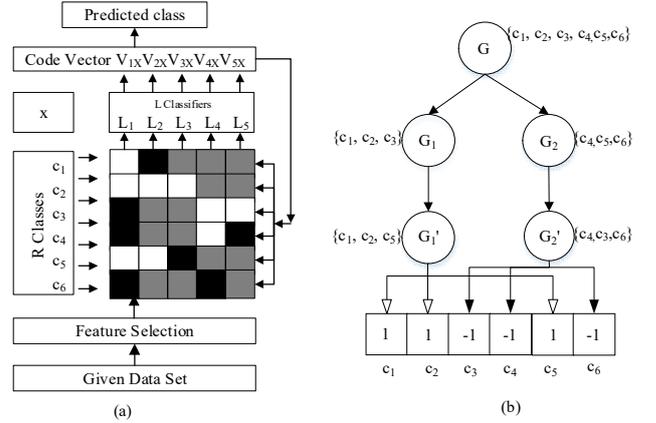

Fig. 1. (a) the workflow of algorithm; (b) an example of encoding of ECOCECS

```
Input: G={c1, c2, …, cR};
Output: M
1:  G is randomly divided into two groups, G1 and G2;
2:  I ←DC index between G1 and G2;
3:  for i in range [1,2]
4:      for each class in Gi
5:          Calculate D_{k,i} on formula (4) or (5);
6:          a_i ← argmax(D_{k,i});
                    k
7:  G'1/G'2 ← exchange a1 and a2 in G1/ G2;
8:  I'← new DC index between G'1 and G'2;
9:  if I' is smaller than I
10:     Replace I, G1 and G2 with I', G'1 and G'2;
11:     go to step 3-8;
12: else
13:     Add a new column to M based on assignment scheme
        in G1 and G2;
14: for i in range [1,2]
15:     if Gi contains one more classes
16:         G ← G_i and go to step 1;
17:     else
18:         Return M;
```

Fig. 2. Pseudo code of ECOCECS

In the algorithm, G is divided into two groups $G_1$ and $G_2$ with equal size at random firstly. By labelling samples in $G_1$ and

$G_2$ as +1 and -1, DC index I is calculated by formula (1) for N2 or formula (2-3) for N3. Then class complexity of each class is evaluated by formula (4) or (5). The most complex classes $a_1$ and $a_2$ in $G_1$ and $G_2$ are exchanged to form two new groups, $G_1$' and $G_2$'. Then the new index I' for $G_1$',$G_2$' is calculated by formula (1) or (2). If I' is lower than I, $G_1$, $G_2$ and I would be replaced. This exchange process ends when I can't be lower any more. Then the class assignment scheme of $G_1$ and $G_2$ is recorded as a new column, added to M. This process can be illustrated as a binary tree, and an example is shown in Fig.1(b). This algorithm runs iteratively until both $G_1$ and $G_2$ contain only one class. After this algorithm stops, a coding matrix is generated. As N2 and N3 exchange classes based on different principles, they would produce diverse coding matrices.

## IV. EXPERIMENTS AND DISCUSSIONS

### A. Experiment Settings

To validate the proposed algorithm, experiments are carried out on five multiclass datasets (as listed in Table 1). SVM and NativeBayes(NB) are used as base classifiers, and three filter feature selection methods ROC, T-test and Wilcoxon are deployed. In addition, One VS One (OVO), One VS All (OVA), Ordinal ECOC(Ordinal), DECOC, ECOCONE and ECOC-Forest(Forest) methods are used for comparison. The last three ECOC methods are provided by ECOC library[5], and the rest are supported by MATLAB 2016 toolboxes. Default settings of all algorithms are adopted.

TABLE I. SUMMARY OF DATASETS

| # | #Dataset | #classes | #features | #training samples | #test samples | Ref. |
|---|---|---|---|---|---|---|
| 1 | Breast | 5 | 9216 | 54 | 30 | [12] |
| 2 | Cancers | 11 | 12,533 | 86 | 74 | [13] |
| 3 | DLBCL | 6 | 4026 | 58 | 30 | [14] |
| 4 | Leukemia | 3 | 12582 | 57 | 15 | [15] |
| 5 | Lung | 3 | 7129 | 64 | 32 | [16] |

Fscore and accuracy are often used to evaluate and compare the performances of different algorithms. The original Fscore and accuracy are designed for binary problems. When applied in a multiclass problem, the average Fscore and accuracy among classes are used. That is, for the i-th binary problem, positive rate ($P_i$), negative rate ($N_i$), true positive ($TP_i$), true negative ($TN_i$), false positive ($FP_i$) and false negative ($FN_i$) are calculated by means of OVA. In this way, the i-th class is regard as the positive class, and others are labelled as the negative class. The final score is the average of all binary problems, as shown by formulas (6-9). Here β is set to 1 to get balanced results.

$$\text{Accuracy} = \text{avg}(\sum_{i=1}^{R} \frac{TP_i + TN_i}{P_i + N_i}) \quad (6)$$

$$\text{Precision} = \text{avg}(\sum_{i=1}^{R} \frac{TP_i}{TP_i + FP_i}) \quad (7)$$

$$\text{Recall} = \text{avg}(\sum_{i=1}^{R} \frac{TP_i}{P_i}) \quad (8)$$

$$\text{Fscore} = \text{avg}(\sum_{i=1}^{R} \frac{(\beta^2+1)*Presicion_i*Recall_i}{\beta^2*Presicion_i+Recall_i}) \quad (9)$$

### B. Experiment results

1) *the analysis of Local improvement algorithms*

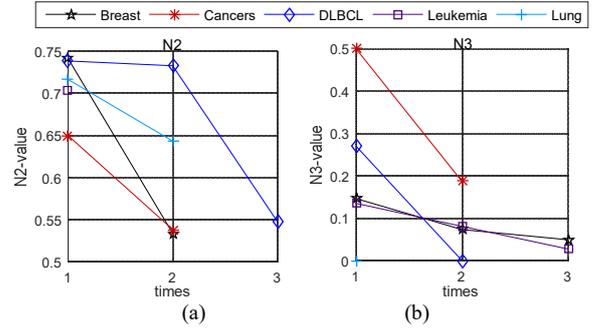

Fig. 3. the change of complexity using Wilcoxon and feature size is 80.

Fig.3 describes the change of complexity indices over all datasets in a class adjustment process. When the original data sets appear high overlapping situation, especially the DLBCL dataset, the class exchange process is of higher probability to occur multiple times. As a result, for N2 based ECOCECS, the index drops to 0.55 after twice exchanges, only about 2/3 of the original. And the N3 value on DLBCL drops to 0 after an exchange. And the similar results obtained by N2 and N3 on other data sets, showing that our algorithm can effectively optimize the class assignment schemes.

2) *Analysis of accuracy obtained by ECOC methods*

Table2 lists the accuracies and Fscore values by different approaches with top 80 features. It is found that accuracies obtained by ECOCECS based schemes are usually higher compared to other ECOC methods. The results obtained by N2 based ECOCECS method are 11% higher than that by OVA and Forest-ECOC. On the Breast data set, the accuracy rate N2 algorithm is 49% higher than OVA, 30% higher than the traditional ECOC, 25% higher than ECOCONE and 54% more than Forest-ECOC. The results show that the algorithm with N2 could balance the precision and recall situation for dataset, performing more precisely. In addition, N2 based ECOCECS accomplishes the highest average datasets accuracy. At the same time, the results of Fscore are similar to those of the accuracy results, revealing that N2 based ECOCECS can produce more balanced results.

Although the N2 based approach wins most cases, the N3 algorithm just wins on the Leukemia data, achieving close performance to other algorithms. However, it should be noted that only our algorithm and OVA, DECOC require R-1 base learners, and other algorithm need much more base learners. So the N3 based algorithm can obtain similar results by a much smaller ensemble scale. While N2 based algorithm is more robust and effective compared with other state-of-art ECOC approaches. Similar conclusions can be drawn from Table3.

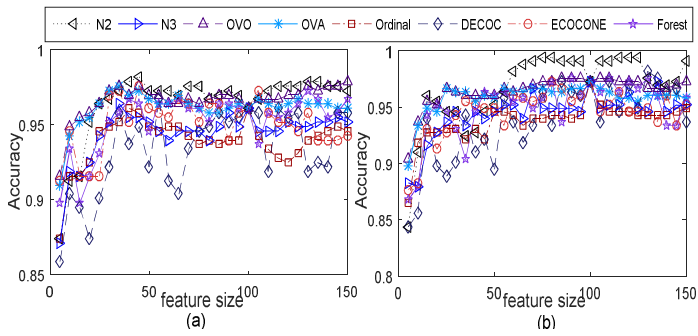

Fig. 4. The fluctuates of accuracy rates over Cancers dataset with various feature size when using NB learner with two FS methods: (a) ROC; (b) Wilcoxon.

Fig.4 shows that the accuracies obtained by eight ECOC methods change with varying feature size over Cancers data set using NB. It is found that when features size is larger than 50, most of ECOC algorithms reached 90% accuracy, depicting that almost all multi-class classification issues still require a mass of genes to ensure good accuracy. Among the results, our algorithm always maintains a lead position (97%-99%). In additional, it is clear that the results corresponding to N2 and N3 algorithms fluctuate slighter compared to and DECOC and ECOCONE algorithms with drastic change. In fact, the performances of DECOC and ECOCONE are susceptible to input data, while there are no clear relationships between the feature size and classification effectiveness for ECOCECS. Furthermore, features picked up by discrepant methods make small influence for classification performance. So our methods are efficient and roust for solving the multi-classification problem.

## V. CONCLUSION

In this paper, we propose ECOCECS algorithms based on DC measures, including N2 and N3 measures, to enhance performance of current ECOC algorithm, and two search methods based on N2 and N3 measure principles are also created to decrease data overlapping region. The main idea of ECOCECS is to utilize the DC measures to form a binary tree from top to bottom, then each node of the binary tree is coded mapped as code matrix. When the parent node of binary tree is divided into child nodes, the sub-node allocation is adjusted with the local search algorithms to guarantee the maximum separability between the child nodes. In experiments, ROC and Wilcoxon methods are used to filter five multi-class microarray datasets. And six prominent ECOC algorithms are deployed for comparisons. The results show that our algorithms win all almost all cases. The results prove that our approaches perform much better than other ECOC methods with great balance of accuracy and Fscores.

TABLE II. ECOC METHODS RESULTS OBTAINED BY NB AND WILCOXON WITH 80 FEATURES

| Methods | Breast | | Cancers | | DLBCL | | Leukemia | | Lung | | Average | |
|---|---|---|---|---|---|---|---|---|---|---|---|---|
| | *Accuracy* | *Fscore* | *Accuracy* | *Fscore* | *Accuracy* | *Fscore* | *Accuracy* | *Fscore* | *Accuracy* | *Fscore* | *Accuracy* | *Fscore* |
| N2-ECOCECS | **0.99** | **0.97** | **0.99** | **0.74** | 0.98 | 0.90 | **1.00** | **1.00** | 0.90 | 0.80 | **0.97** | **0.88** |
| N3-ECOCECS | 0.89 | 0.63 | 0.95 | 0.53 | 0.92 | 0.58 | **1.00** | **1.00** | **0.94** | 0.88 | 0.94 | 0.72 |
| OVO | 0.96 | 0.89 | 0.97 | 0.58 | **1.00** | **1.00** | **1.00** | **1.00** | 0.92 | 0.85 | **0.97** | 0.86 |
| OVA | 0.89 | 0.48 | 0.97 | 0.60 | 0.94 | 0.68 | **1.00** | **1.00** | 0.90 | 0.80 | 0.94 | 0.71 |
| Ordinal | 0.92 | 0.67 | 0.96 | 0.55 | 0.97 | 0.79 | **1.00** | **1.00** | 0.90 | 0.80 | 0.95 | 0.76 |
| DECOC | 0.95 | 0.83 | 0.94 | 0.50 | 0.97 | 0.86 | **1.00** | **1.00** | **0.94** | **0.91** | 0.96 | 0.82 |
| ECOCONE | 0.93 | 0.72 | 0.97 | 0.69 | 0.97 | 0.79 | **1.00** | **1.00** | 0.92 | 0.85 | 0.96 | 0.81 |
| Forest | 0.88 | 0.43 | 0.93 | 0.48 | 0.92 | 0.55 | **1.00** | **1.00** | 0.90 | 0.80 | 0.93 | 0.65 |

TABLE III. ECOC METHODS RESULTS OBTAINED BY SVM AND WILCOXON WHEN 80 FEATURES

| Methods | Breast | | Cancers | | DLBCL | | Leukemia | | Lung | | Average | |
|---|---|---|---|---|---|---|---|---|---|---|---|---|
| | *Accuracy* | *Fscore* | *Accuracy* | *Fscore* | *Accuracy* | *Fscore* | *Accuracy* | *Fscore* | *Accuracy* | *Fscore* | *Accuracy* | *Fscore* |
| N2 | **0.99** | **0.97** | **0.99** | 0.66 | 0.94 | 0.63 | **1.00** | **1.00** | 0.92 | 0.86 | **0.97** | 0.82 |
| N3 | 0.91 | 0.69 | 0.95 | 0.53 | 0.92 | 0.55 | **1.00** | **1.00** | 0.90 | 0.83 | 0.93 | 0.72 |
| OVO | 0.96 | 0.89 | 0.98 | 0.60 | **0.98** | **0.82** | **1.00** | **1.00** | 0.90 | 0.83 | 0.96 | **0.83** |
| OVA | 0.89 | 0.48 | 0.98 | 0.66 | 0.94 | 0.68 | **1.00** | **1.00** | 0.90 | 0.83 | 0.94 | 0.73 |
| Ordinal | 0.92 | 0.67 | 0.97 | 0.57 | 0.97 | 0.79 | **1.00** | **1.00** | 0.90 | 0.83 | 0.95 | 0.78 |
| DECOC | 0.93 | 0.83 | 0.95 | 0.51 | 0.96 | 0.77 | **1.00** | **1.00** | 0.90 | 0.83 | 0.95 | 0.79 |
| ECOCONE | 0.91 | 0.70 | 0.98 | **0.70** | 0.97 | 0.79 | **1.00** | **1.00** | 0.90 | 0.83 | 0.95 | 0.81 |
| Forest | 0.92 | 0.82 | 0.96 | 0.57 | 0.92 | 0.55 | **1.00** | **1.00** | **0.92** | 0.86 | 0.94 | 0.76 |

ACKNOWLEDGMENT

This work is supported by National Key Technology Research and Development Program of the Ministry of Science and Technology of China (2015BAH55F05); Natural Science Foundation of Fujian Province (No. 2016J01320 and 2015J05129), and National Natural Science Foundation of China (Grant No.61502402 and 61772023).

REFERENCES

[1] V. Bolón-Canedo, N. Sánchez-Maroño, A. Alonso-Betanzos, J. M. Benítez, and F. Herrera, "A review of microarray datasets and applied feature selection methods," Information Sciences An International Journal, vol. 282, no. 5, pp. 111-135, 2014.

[2] D. B. Allison, X. Cui, G. P. Page, and M. Sabripour, "Microarray data analysis: from disarray to consolidation and consensus," Nature Reviews Genetics, vol. 7, no. 1, p. 55, 2006.

[3] K. H. Liu, Z. H. Zeng, and V. T. Y. Ng, "A Hierarchical Ensemble of ECOC for Cancer Classification Based on Multi-Class Microarray Data," Information Sciences, vol. 349, pp. 102-118, 2016.

[4] J. Zhou and C. Y. Suen, "Unconstrained numeral pair recognition using enhanced error correcting output coding: a holistic approach," in Document Analysis and Recognition, 2005. Proceedings. Eighth International Conference on, 2005, pp. 484-488 Vol. 1.

[5] S. Escalera, O. Pujol, and P. Radeva, "Error-Correcting Ouput Codes Library," Journal of Machine Learning Research, vol. 11, no. 1, pp. 661-664, 2010.

[6] N. Arvanitopoulos, D. Bouzas, and A. Tefas, "Subclass Error Correcting Output Codes Using Fisher's Linear Discriminant Ratio," in International Conference on Pattern Recognition, 2010, pp. 2953-2956.

[7] T. K. Ho and M. Basu, "Complexity Measures of Supervised Classification Problems," IEEE Transactions on Pattern Analysis & Machine Intelligence, vol. 24, no. 3, pp. 289-300, 2002.

[8] E. B. Mansilla and T. K. Ho, "On Classifier Domains of Competence," in International Conference on Pattern Recognition, 2004, pp. 136-139 Vol.1.

[9] L. Li, "Data complexity in machine learning and novel classification algorithms," 2006.

[10] J. Cano and -Ram, "Analysis of data complexity measures for classification," Expert Systems with Applications An International Journal, vol. 40, no. 12, pp. 4820-4831, 2013.

[11] N. Anwar, G. Jones, and S. Ganesh, "Measurement of data complexity for classification problems with unbalanced data," Statistical Analysis & Data Mining, vol. 7, no. 3, pp. 194-211, 2014.

[12] C. M. Perou et al., "Molecular portraits of human breast tumours," Nature, vol. 490, no. 7418, p. 61, 2000.

[13] A. I. Su et al., "Molecular Classification of Human Carcinomas by Use of Gene Expression Signatures," Cancer Research, vol. 61, no. 20, p. 7388, 2001.

[14] M. A. Shipp et al., "Diffuse large B-cell lymphoma outcome prediction by gene-expression profiling and supervised machine learning," vol. 8, no. 1, pp. 68-74, 2002.

[15] S. A. Armstrong et al., "The MLL dependent gene expression profile: Characterization of a unique leukemia and identification of a potential molecular therapeutic target," in Blood, 2001, pp. 800A-800A.

[16] G. J. Gordon et al., "Translation of Microarray Data into Clinically Relevant Cancer Diagnostic Tests Using Gene Expression Ratios in Lung Cancer and Mesothelioma," Cancer Research, vol. 62, no. 17, pp. 4963-7, 2002.